\documentclass{article}

\usepackage{arxiv}

\usepackage{graphicx}
\usepackage{booktabs}
\usepackage{subfig}
\usepackage{amsmath}
\usepackage{amssymb}
\usepackage{float}
\usepackage[hidelinks]{hyperref}
\usepackage{color}
\usepackage{caption}

\renewcommand{\headeright}{Accepted in \textit{Physica A}}
\renewcommand{\undertitle}{Accepted in \textit{Physica A: Statistical Mechanics and its Applications}}
\renewcommand{\shorttitle}{Multiple Descents as Order-Chaos Transitions in LSTM}

\title{Multiple Descents in Deep Learning as a Sequence of\\Order-Chaos Transitions in LSTM Networks}

\author{
  Wenbo Wei \\
  Department of Physics \\
  National University of Singapore, 117551, Singapore \\
  \And
  Fan Xu \\
  Department of Physics \\
  National University of Singapore, 117551, Singapore \\
  \And
  Nicholas Jia Le Chong \\
  Department of Physics \\
  National University of Singapore, 117551, Singapore \\
  \And
  Choy Heng Lai \\
  Department of Physics \\
  National University of Singapore, 117551, Singapore \\
  \AND
  Ling Feng\thanks{Corresponding author: \href{mailto:0@criticality.ai}{\texttt{0@criticality.ai}}} \\
  Department of Physics, National University of Singapore, 117551, Singapore \\
  Institute of High Performance Computing (IHPC), A*STAR, 138632, Singapore \\
}

\date{}

\begin{document}

\maketitle

\begin{abstract}
We observe a novel `multiple-descent' phenomenon during the learning process of a recurrent neural network called long-short-term memory (LSTM) networks during its training on real-world task, in which the performance goes through long cycles of up and down trends multiple times after the model is overtrained.
By carrying out asymptotic stability analysis of the models, we found that the cycles in performance---indicated by loss function in test data---are closely associated with the phase transition process between order and chaos of the model, and the local optimal training step are consistently at the critical transition point between the two phases.
More importantly, the most optimal point of the model usually occurs at the first transition from order to chaos, where the `width' of the `edge of chaos' is often the widest, allowing the best exploration of weight configurations for learning.
\end{abstract}


\section{Introduction}
\label{sec:level1}

In deep learning, understanding the training dynamics has become paramount for enhancing model performance, generalization, and robustness. The training of deep neural networks involves navigating through complex, high-dimensional parameter spaces, where the interplay between model complexity, dataset characteristics, and learning algorithms dictates the learning trajectory. This process is far from straightforward, often characterized by phenomena such as overfitting, underfitting, and various forms of swings in performance metrics.

%\subsection{Importance of Understanding Training Dynamics}

The dynamics of training deep neural networks are critical for several reasons. Generalization is a primary concern in machine learning, focusing on the model's ability to generalize from training data to unseen data. Understanding how training dynamics affect generalization can lead to better strategies for model tuning \cite{Geiger2020, Zhang2021}. Optimization within deep networks is challenging due to the non-convex nature of loss landscapes, where insights into training dynamics inform the choice of optimizers, learning rates, and regularization techniques \cite{Reddi2018}. Overfitting and underfitting are pivotal issues; the former occurs when a model memorizes training data too closely, while the latter happens when it fails to capture underlying patterns. Studying training dynamics aids in determining when to stop training to avoid these pitfalls \cite{Srivastava2014, Goodfellow2016}. Lastly, dissecting the training process can enhance model interpretability by revealing which parts of the networks are crucial for learning \cite{Zeiler2014, Olah2017}.

%\subsection{Current Status in Training Dynamics}

The current understanding of training dynamics in deep learning involves several key concepts. The concept of the edge of stability posits that networks perform best when their parameters balance between order and chaos, optimizing learning capacity \cite{Cohen2021, Arora2018}. The Neural Tangent Kernel (NTK) framework offers insights into how wide neural networks learn by resembling kernel methods in the infinite-width limit \cite{Jacot2018, Lee2019}. Implicit regularization by optimization algorithms leads to better-than-expected generalization \cite{Neyshabur2017, Gunasekar2018}. Exploring the loss landscape provides a visual understanding of how different optimization paths lead to minima with varied generalization capabilities \cite{Garipov2018}. The scale of gradient noise during training correlates with model generalization, with higher noise often linked to better generalization \cite{Smith2018, Keskar2017}. Lastly, normalization techniques like batch normalization have shown to stabilize learning and improve generalization \cite{Ioffe2015, Santurkar2018}.

The double descent phenomenon has emerged as a significant area of study in recent years, particularly in the context of deep learning. Initially observed in classical statistics \cite{Belkin2019}, this phenomenon describes a U-shaped curve in model performance where, after an initial decrease, the error on the test dataset increases with model complexity or training duration before decreasing again. This behavior challenges the traditional understanding of how the trade-off between model complexity and generalization works, which is often described by the
bias-variance trade-off and has been both empirically observed and theoretically analyzed in the context of neural networks \cite{Nakkiran2021, Mei2022}. %However, while double descent provides insights into how model complexity affects generalization, several open questions remain. These include the exact mechanisms behind why increasing model capacity can lead to better performance after an initial decline, the role of dataset size and noise, and how these observations extend to different types of architectures or learning tasks.

Our study introduces a novel observation of `multiple descents' in the training of deep learning models, specifically in Long Short-Term Memory (LSTM) networks~\cite{hochreiter1997long}. LSTM network is a type of recurrent neural network (RNN) that is widely used for processing sequential data, such as natural language, time series, and speech. It is designed to capture long-term dependencies in data through its unique architecture of gates and memory cells, and has been successfully applied in various applications, including language modeling, machine translation, and sentiment analysis. Its working principle is to first convert the input tokens (like words in English) into numerical values through a mathematical transformation called embedding layer, then process the sequence of embedding vectors one at a time through a complex neural network module called the LSTM layer (as a series of mathematical operations described below) to capture temporal dependencies, and finally use a fully connected network layer to produce a binary value of 0 or 1, which is the output value of the whole neural network. In our task, 0 means the input movie review is a negative review and 1 means it is a positive review. The LSTM layer can be mathematically described below:
\begin{equation}
    \label{eq:lstm_cell}
    \begin{aligned}
        \boldsymbol{i}_t &= \sigma_s\left(\boldsymbol{W}_i\boldsymbol{x}_t + \boldsymbol{U}_i\boldsymbol{h}_{t-1} + \boldsymbol{b}_i\right), \\
        \boldsymbol{f}_t &= \sigma_s\left(\boldsymbol{W}_f\boldsymbol{x}_t + \boldsymbol{U}_f\boldsymbol{h}_{t-1} + \boldsymbol{b}_f\right), \\
        \boldsymbol{o}_t &= \sigma_s\left(\boldsymbol{W}_o\boldsymbol{x}_t + \boldsymbol{U}_o\boldsymbol{h}_{t-1} + \boldsymbol{b}_o\right), \\
        \tilde{\boldsymbol{c}_t} &= \sigma_t\left(\boldsymbol{W}_c\boldsymbol{x}_t + \boldsymbol{U}_c\boldsymbol{h}_{t-1} + \boldsymbol{b}_c\right), \\
        \boldsymbol{c}_t &= \boldsymbol{f}_t \odot \boldsymbol{c}_{t-1} + \boldsymbol{i}_t \odot \tilde{\boldsymbol{c}_t}, \\
        \boldsymbol{c}_{\mathrm{out}t} &= \sigma_t\left(\boldsymbol{c}_t\right), \\
        \boldsymbol{h}_t &= \boldsymbol{o}_t \odot \boldsymbol{c}_{\mathrm{out}t}, \\
    \end{aligned}
\end{equation}
where $\boldsymbol{x}_t$ is the input embedding vector from a word token at position $t$ in the movie review; $\sigma_s$ is the sigmoid activation function; $\sigma_t$ is the hyperbolic tangent activation function; $\odot$ denotes element-wise multiplication; and $\boldsymbol{W}$, $\boldsymbol{U}$, and $\boldsymbol{b}$ are the weight matrices and bias vectors for the respective gates and cell state. Each of the other variables are simply different intermediate vectors in the LSTM cell to process the input data, and do not carry any physical meaning. For our experiment, we focus on the output vector $\boldsymbol{h}_t$ as the main variable to analyze the order-chaos transitions, and we will refer to it as the output recurrent unit of the LSTM cell for short.

We explore this through an extensive analysis of LSTM networks trained on the Large Movie Review Dataset for sentiment analysis during each time step of the training process, i.e. training epoch. Specifically, one can treat the LSTM as a black box non-linear dynamical system, and use the tools in physics such as Lyapunov exponent or asymptotic stability analysis to extract the order/chaos properties of the model at each epoch, and use the test loss as a measure of the model performance. It is then illucinating to explore the relationship between the dynamical phases and the model performance when the model is trying to learn the data.
 Our research reveals that during the training process, particularly in the overfitting phase, the test loss exhibits multiple cycles of increases followed by a sharp decline, a pattern not fully captured by existing models like double descent. While multiple descent phenomena have been observed in other contexts such as random feature models \cite{meng2024multiple} and unsupervised autoencoders \cite{rahimi2024unveiling}, our work provides unique insights into the relationship between multiple descents and order-chaos transitions in recurrent networks. And more importantly, the most optimal epoch is consistently at the first transition from order to chaos, where the `edge of chaos' is the widest, allowing the best exploration of better weight configurations for learning.

One core contribution of this study is the identification of these cycles with specific phase transitions. We propose that each descent in test loss corresponds to a transition between order and chaos state during the evolution of the neural network from training, and the optimal performance is achieved at the edge of these transitions. By employing techniques from dynamical systems, specifically asymptotic stability analysis, we measure how perturbations in initial conditions propagate through the network, offering insights into when these order-chaos transitions occur.

We will detail the experimental setup where we over-train an LSTM model to observe these dynamics, provide some theoretical explanation why these transitions might occur, by drawing parallels with known behaviors in non-linear systems like the $\tanh$ map, and discuss the implications of these findings for training deep learning models, particularly in terms of finding optimal training epochs.

\section{Experimental Setup}

We build a basic LSTM model to perform sentiment analysis of the Large Movie Review Dataset \cite{dataset}, which contains 50,000 labelled movie reviews from the Internet Movie Database (IMDb) and is a popular open-source dataset for natural language processing. The model is over-trained to 1,000 epochs to induce overfitting in order to explore the rich order chaos transition behaviors. The embedding layer projects the individual words (tokens) into a learned continuous vector space, the LSTM layer performs the time-series analysis, and the output layer performs the final classification. The hyperparameters of the model are summarized in Table~\ref{tab:table1}.
We partition 70\% of the dataset for training and 30\% for testing, padding/truncating each review to a fixed length of 500 tokens. The model is optimized with the \texttt{Adam} optimizer with a learning rate of 0.0005.

\begin{table}[tb]
\caption{Hyper-parameter for the structure of the model}
\label{tab:table1}
\centering
\begin{tabular}{|l|c|}
\hline
\textbf{Layer} & \textbf{Output Dimension} \\
\hline
Embedding Layer & 32 \\
LSTM Layer & 60 \\
Fully Connected Layer & 1 \\
\hline
\end{tabular}
\end{table}

\subsection{Asymptotic stability analysis}

We calculate the asymptotic trajectory separation/difference of the output recurrent unit of LSTM cell $\boldsymbol{h}_t$ under perturbation to extract the asymptotic stability of the LSTM model, specifically order and chaos phases. The key principle is the following: suppose we have a dynamical system (in this case the LSTM cell) with the recurrence relation:
\begin{eqnarray}
\boldsymbol k_t=\boldsymbol F(\boldsymbol k_{t-1})
\end{eqnarray}
where $\boldsymbol k_t$ is the state vector of the system at time step $t$, and $\boldsymbol F$ is the non-linear function. In the case of LSTM, $\boldsymbol k_t$ corresponds to the output recurrent unit $\boldsymbol h_t$ at time step $t$, and $\boldsymbol F$ corresponds to the LSTM cell described by Equation~\ref{eq:lstm_cell}.
We add a small noise perturbation $\boldsymbol{\varepsilon}$ drawn from a Gaussian distribution at the 0th timestep to $\boldsymbol{k_0}$, giving another input $\boldsymbol{k_0'}=\boldsymbol{k_0}+\boldsymbol{\varepsilon}$. Iterating both $\boldsymbol{k_0}$ and $\boldsymbol{k_0'}$ for $T$ steps (where $T$ is large) gives two asymptotic values: $\boldsymbol{k_T}$ and $\boldsymbol{k_T'}$. If the system is in an ordered state, the asymptotic distance:
\begin{equation}
D=\vert \boldsymbol k_T'- \boldsymbol k_T\vert
\label{eq:asymptotic_distance}
\end{equation}
converges to 0. Otherwise, they diverge, and it means the system is in the asymptotic chaos phase. Unlike the basic feedforward neural networks, the LSTM network has two different types of inputs. The first is an external input $\boldsymbol{x_t}$ which corresponds to the input token at time step $t$, and the second is the recurrent input $h_{t-1}$ from the previous time step.

\begin{figure}[H]
\centering
\includegraphics[width=1\columnwidth]{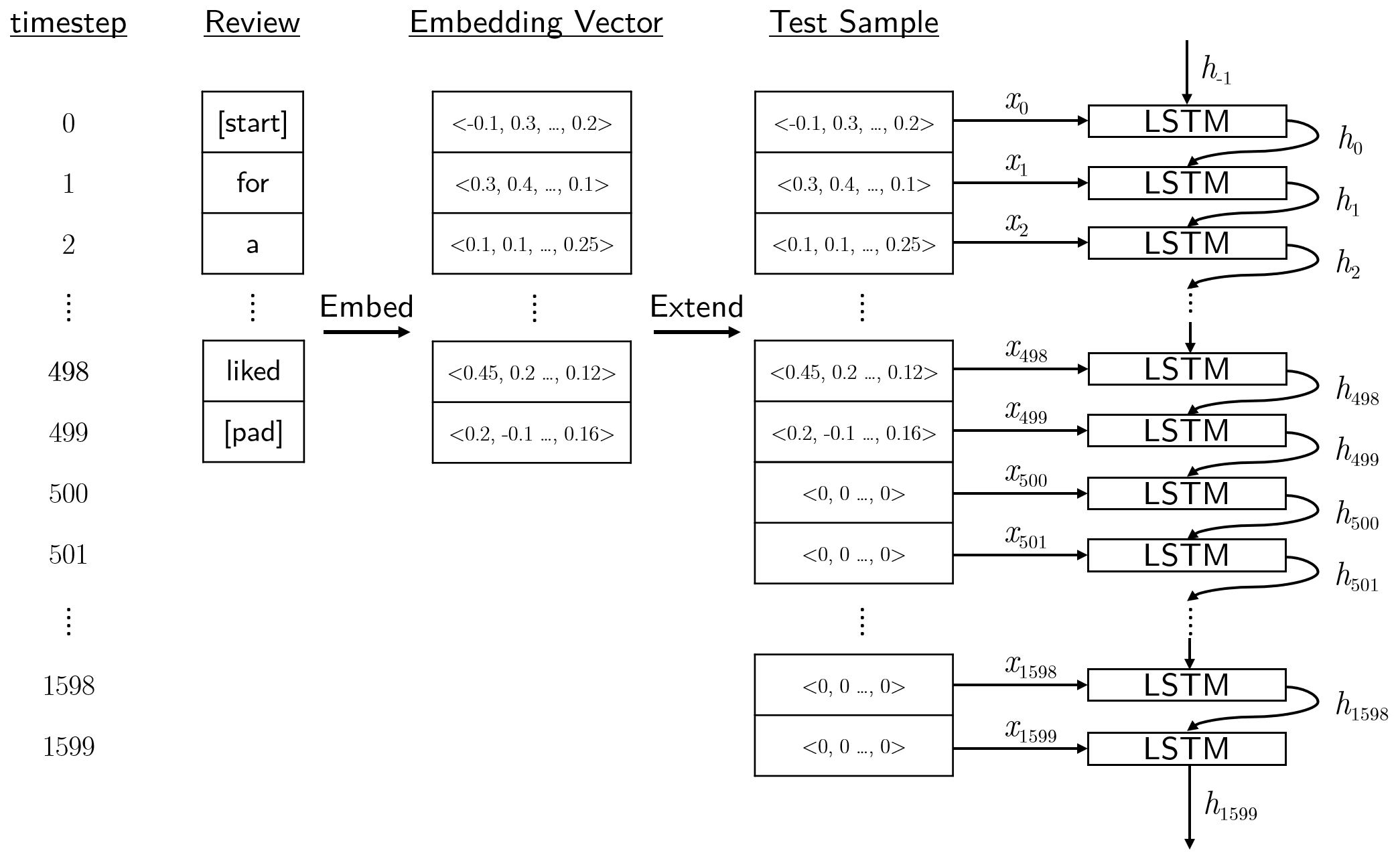}
\caption{{\bf Illustration of our methodology to iterate the LSTM cell to get the asymptotic value of output neuron $\boldsymbol{h_T}$, where $T$ is large at $T=1599$.} The first 500 iterations use words from the movie reviews as input, after which only $\boldsymbol{x_{t\ge 500}}=\boldsymbol{0}$ vectors are used as LSTM cell inputs to extract the order/chaos properties of the model. Note that to isolate the dynamical process to focus only on the LSTM cell, we freeze the embedding layer that transforms the input tokens to embedding vectors during the entire training process, and we also freeze the final output layer after the LSTM cell.}
\label{fig:method}
\end{figure}

The intrinsic order/chaos property of the network is to be characterized when there is no continuous external inputs or driving force to it. Therefore, the first 500 iterations of the LSTM are ignored for asymptotic analysis, as they are fed with non-zero tokens from the words in the reviews. After 500 iterations (input review length with padding), we let the LSTM continue running or iterating without external input values, i.e., $\boldsymbol x_{t>500}=0$. To get the asymptotic distance, a noise perturbation is introduced to the output unit $\boldsymbol h_{500}$ right after the final token from the input reviews $\boldsymbol{x_{499}}$,  i.e., $\boldsymbol h'_{500}=\boldsymbol h_{500}+\boldsymbol{\epsilon}$. Next, we evaluate the asymptotic trajectory difference between the original and perturbed output values of the LSTM cell, i.e., $|\boldsymbol h'_T- \boldsymbol h_T|$, with large $T$ value at $T=1599$.

\subsection{Detailed method}

Due to the high computational cost of our operation, we randomly sample only 500 reviews from the test dataset, which we then use across the rest of the experiments. This subset was used to keep our experiment consistent while being computationally tractable.

After each epoch during the training process, we obtain 500 asymptotic distances $D$ in equation~\ref{eq:asymptotic_distance} from the 500 reviews, and then evaluate the average of their $\ln$ values to extract the order/chaos state of the model at that epoch. Note that to account for machine precision in the numerical process, we add a tiny constant value $\exp(-15)$ to each asymptotic distance and use it as a reference to 0. This also means we regard $\ln D=-15$ to be the case where the model is in the ordered phase, as any value lower than may be due to machine precision and can be effectively treated as 0.

Fig.~\ref{fig:method} demonstrates our main methodology for measuring the asymptotic distances. During each epoch of the training process, our steps can be summarized as follows:

\begin{enumerate}
\item Pass a review from the test sample through the embedding layer to obtain $\boldsymbol x_0\cdots \boldsymbol x_{499}$, the embedding vectors from the real review. Extend the array of embedding vectors to 1600 time steps with the zero vectors $\boldsymbol x_{500}\cdots \boldsymbol x_{1599}$.
\item At each timestep, propagate the review vector $\boldsymbol x_t$ along with the hidden vector $\boldsymbol h_{-1}$ through the layer to the last timestep (in our case 1600) to obtain $\boldsymbol h_{1599}$.
\item Repeat steps 1 and 2 with the same review, with the exception of adding a small noise to the hidden unit before the zero input vectors at step 500, i.e. $\boldsymbol h'_{500}=\boldsymbol h_{500}+\boldsymbol\varepsilon$ (where $\boldsymbol\varepsilon$ is $d$-dimensional Gaussian noise), obtaining $\boldsymbol h'_{1599}$.
\item Calculate the distance between $\boldsymbol h'_{1599}$ and $\boldsymbol h_{1599}$, giving the asymptotic distance for the $i$-th review $D_{i}=\vert \boldsymbol h'_{1599}-\boldsymbol h_{1599}\vert$. Finally, add $\exp(-15)$ giving $D'_{i}=D_{i}+\exp(-15)$.
\item Repeat steps 1 to 4 for all 500 reviews in our test sample.
\item Calculate the $\ln$ value of the geometric mean $\tilde{D} = \log\left(\sqrt[{500}] {D'_1D'_2\cdots D'_{500}}\right)$, at this epoch for the 500 samples. We call $\tilde{D}$ the `asymptotic distance' for short.
\item Calculate two more quantities to confirm the phase transitions: 1. the reduced sum $\boldsymbol h_{1599} \cdot \boldsymbol{1}$ for each of the 500 reviews, where $\boldsymbol{1}$ is a vector of 1s, to visualize the bifurcation process to chaos of the model; 2. the finite time Lyapunov exponent (FTLE) using the Benettin estimator~\cite{Benettin1980a}.
\item Repeat steps 1 to 7 for every epoch during model training, measuring $\tilde{D}$, $\boldsymbol h_{1599} \cdot \boldsymbol{1}$ {and the FTLE} across the entire training process.
\end{enumerate}

To reiterate, in step 6, if the asymptotic distance $\tilde{D}$ is very negative (minimal value is $-15$ due to machine precision adjustment), it indicates the model is in the ordered phase. If it is significantly larger than $-15$, it indicates the model is in the chaotic phase. To double check the order/chaos phase of the model, in step 7, the reduced sum $\boldsymbol h_{1599} \cdot \boldsymbol{1}$ is calculated to project the high dimensional $\boldsymbol h_{1599}$ to 1 dimension to visualize the order/chaos states of the model.
Finte time Lyapunove exponent is also used to further check the phase transitions, as it is a common and robust indicator used in dynamical systems analysis. If the reduced sums of the 500 follow-up reviews converges to the same value, it indicates the model is in the ordered phase. If the reduced sums of the 500 reviews scatter, it indicates the model is in the chaotic phase.
\section{Results on multiple descents and order-chaos transitions}
\label{section:experiment}
\begin{figure}[!ht]
\centering
\includegraphics[width=0.75\columnwidth]{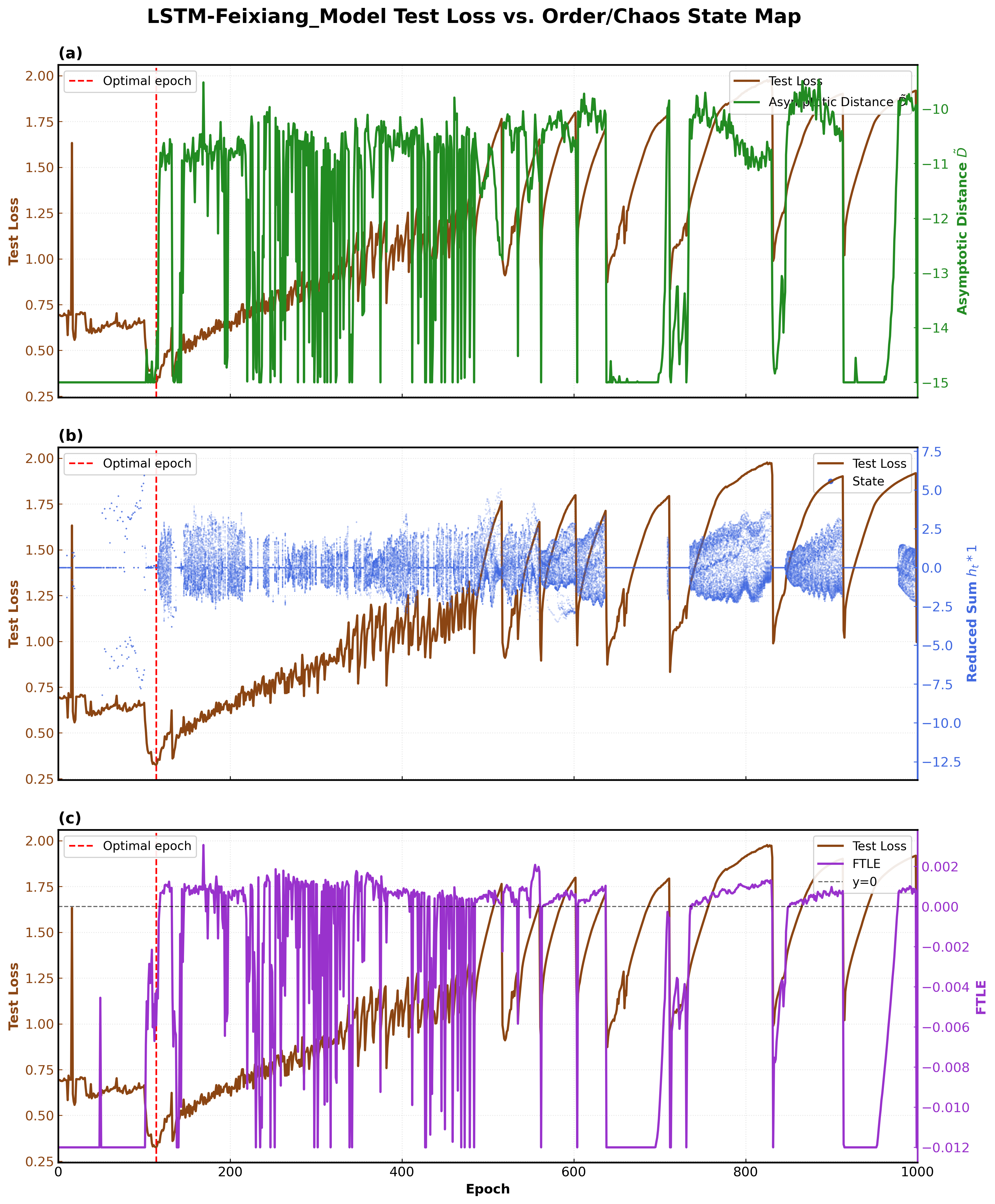}
\caption{{\bf Multiple descents through a sequence of order chaos transitions during the training process of LSTM.} The dynamical process of converging towards this figure is in Movie 1. (a) The average asymptotic log distances $\tilde{D}$ (green) under perturbation is used to indicate order/chaos states. The optimal epoch of LSTM-1 is 114 with an accuracy of 88.34\%. Multiple descents are seen in the overfitting regime at epochs $>450$. When the asymptotic distance is at $-15$, it means two slightly different initial input values will converge to the same value at long enough iterations of the LSTM cell, indicating order phase. If the asymptotic distance is large, it means the model is at chaotic phase. (b) (Please refer to Movie 1 for a clearer visualization.) The `bifurcation map' (blue) is shown together with the test loss (brown). The `bifurcation map' is drawn by plotting the reduced sums $\boldsymbol h_{1599} \cdot \boldsymbol{1}$ for each of the 500 review samples at every epoch. Note that within every epoch, the average of all 500 reduced sums has been subtracted from each reduced sum value for the ease of visualization. Similarly, if the different samples converge to the same value, it indicates order phase. If the samples spread out, it indicates chaotic phase.
(c), Finite time Lyapunov exponent (FTLE) vs. the epochs, and FTLE=0 is the transition point between order and chaos. It can also be seen the loss descents are happening at the transition epochs when FTLE crosses zero.}
\label{fig:lstm_chaos}
\end{figure}
By running the training process for 1,000 epochs, we obtain two notable features which are evident from Movie 1 as well as Fig.~\ref{fig:lstm_chaos}. The Movie 1 gives a more dynamic view of the asymptotic changes of the model at each epoch, while Fig.~\ref{fig:lstm_chaos} summarizes the results at the final snapshot. We strongly recommend the readers to view the Movie 1 to appreciate the dynamical process of the model training.

First, when the model is overtrained and over-fitting happens after approximately 500 epochs, long cycles of test loss going up happen, and each cycle ends with a abrupt drop within just a single epoch. As seen in Fig.~\ref{fig:lstm_chaos}, 8 such cycles are clearly visible between epoch 500 and 1000. In each cycle, we can see the asymptotic distance of the model increases when test loss increases, and both quantities suffer from the same sudden and drastic drop near epochs like 600, 700, 850, and 1000 at the end of the cycle. This indicates that between epoch 500 and 1000, during each cycle the model performance gets worse as it becomes less stable (i.e., more chaotic), until suddenly the model performance gets better when it suddenly transitions from the chaotic phase to order phase. 

To even further demonstrate the order/chaos phases during the training process, we also plot the reduced sum $\boldsymbol h_{1599} \cdot \boldsymbol{1}$ for each of the 500 samples in every epoch in Fig.~\ref{fig:lstm_chaos}(b). We do this simple dimensionality reduction to 1 dimension to allow the ease of visualization. A detailed examination of the $\boldsymbol{h_{1599}}$ vectors from different samples show that they are identical when their reduced sums are the same. Note that we have normalized the reduced sums by subtracting the average over 500 samples in each epoch for better visualization, without affecting the conclusion of the results. In the epochs of order phase, the 500 samples converges to only one or several values, a typical feature of order phase; In the epochs of chaos phase, the 500 samples spread out vertically and do not converge. One thing to note in the Figure.~\ref{fig:lstm_chaos} is that, there is a sudden (smaller) descent around epoch 650, yet none of the 3 indicators show any changes there. However, if one looks at the Movie 1 from $t=0$ to $t=300$, the phase before and after epoch 650 are clearly different as they have different attractors; As $t$ increases further, both attractors become order. This is a clear sign of transient chaos, which is a common phenomenon in non-linear dynamics, where the system can exhibit chaotic behavior for a finite time before settling into an ordered state. The fact the test loss suffer a sudden descent here means the transient phase also drastic impacts on the model performance.

Secondly, we can observe the first order-to-chaos transition occurs at epoch 114, when the model also achieves the lowest test loss throughout the whole training process. Such transition can be observed both after epoch 114 in Fig.~\ref{fig:lstm_chaos}. The same phenomenon has been observed in feedforward neural networks, that the `edge of chaos' coincides with the optimal performance of the networks \cite{zhang2024asymptotic,16}. The role of chaotic dynamics in recurrent neural networks has been extensively studied in the context of brain modelling \cite{mattera2025chaotic}, and our findings provide additional evidence for the importance of chaos-order transitions in neural computation. However, our experiment here further indicates that although there are multiple descents with each associated with a transition between order and chaos during the training process, the best performance happens when the model enter from order to chaos for the first time.

\begin{figure}[H]
\centering
\includegraphics[width=\columnwidth]{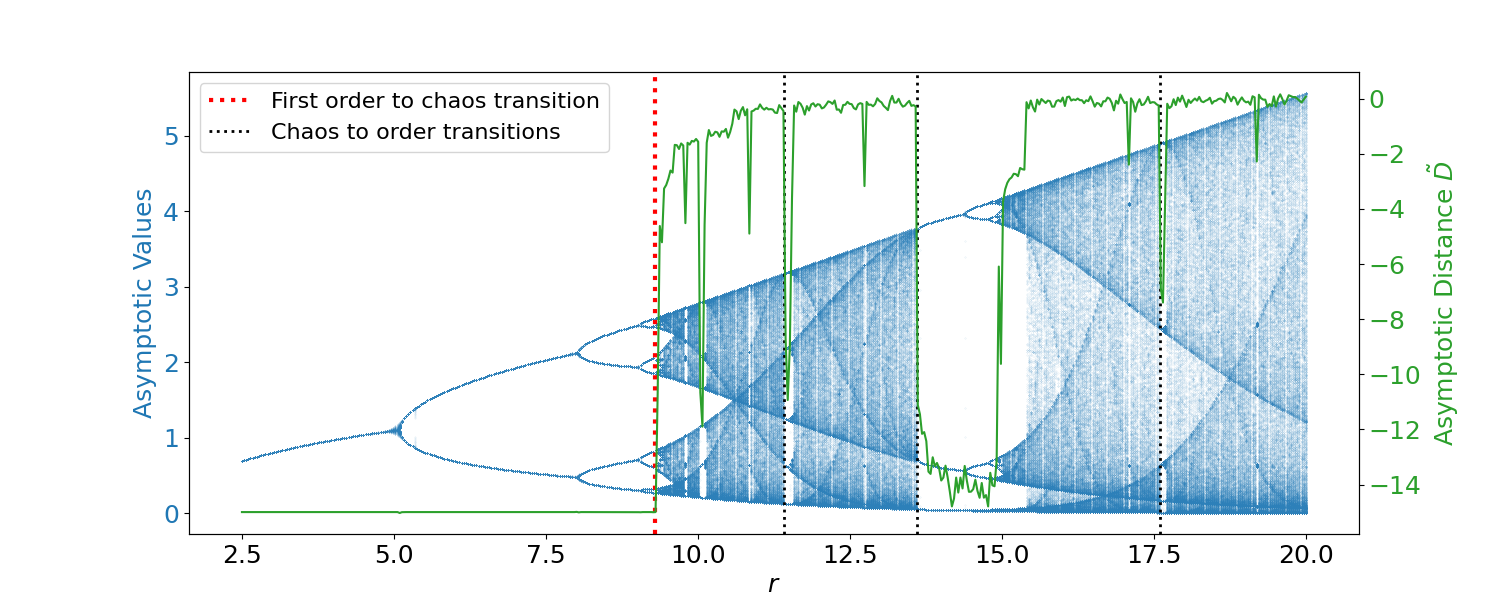}
\caption{{\bf The $\tanh$ bifurcation map in equation (3)}, showing the asymptotic distances (green) and the order/chaos (a.k.a. bifurcation) diagram (blue). 500 random initial values of $k_0$ are used.}
\label{fig:tanh_bifurcation}
\end{figure}

Interestingly, such order chaos transition trends resemble to that of a one-dimensional $\tanh$ map in dynamical systems, or rather the generic non-linear 1D map widely studied in dynamical systems, including the logistic map, at a {\it phenomenological} level. The $\tanh$ map is an extension of the logistic map and is given by the following recurrence relation:
\begin{equation}
k_t = r k_{t-1} (1 - \tanh(k_{t-1}))
\label{eq:tanh_map}
\end{equation}

Since the LSTM equations contain mostly hyperbolic functions as non-linear activations as seen in equation~\ref{eq:lstm_cell}, it is not surprising that LSTM exhibits a similar phase diagram as the $\tanh$ map. And indeed as seen from Fig.~\ref{fig:tanh_bifurcation}, the phase diagram of the $\tanh$ map also exhibits a first order chaos transition at $r\approx 9$, with many subsequent transitions after that. In LSTM during the training process, as we have not applied any regularization, the overall sizes of weight values keep increasing due to stochastic gradient descent(SGD), or more specifically due to the random walk like nature of the SGD. The increasing weights matrix values $\boldsymbol{W}$ in Equation~\ref{eq:lstm_cell} amplifies the linear components Equation~\ref{eq:lstm_cell}. This is similar to the increasing value of amplification factor $r$ in the $\tanh$ map in Equation~\ref{eq:tanh_map}.

It is also widely observed that the first transition from order to chaos in non-linear dynamics like the $\tanh$ map and other non-linear maps is the slowest, as can be also seen in Fig.~\ref{fig:tanh_bifurcation} the wide range of $r$ values below 9. Our results indicate that LSTM has similar behavior, that the first entry from order to chaos is the slowest, followed by many (infinity) other order chaos transitions as the model weights (equivalent to $r$ in the $\tanh$ map) increase overall with training. Since `edge of chaos' is the best weights configuration for the model to process information \cite{16}, that means it is the easiest for the model to explore optimal configuration at the widest transition point, which is exactly the first order chaos transition around epoch 114 in Fig.~\ref{fig:lstm_chaos}. Therefore, the model is the most optimal at that epoch.

One clear evidence that the transition is the widest at the first order to chaos transition in LSTM is that we can observe a wide bifurcation regime in Fig.~\ref{fig:lstm_chaos}(b). From the beginning of training till around epoch 50, we see that the reduced sum of the 500 samples mostly converge to only one unique value, which is typically known as the single fixed point in the order phase. Between epoch 50 and 100, the reduced sum `bifurcates' into two values (one above 0 and one below 0), before chaos kicks in at epoch 114. On the other hand, for the multiple descents regime after epoch 450, the bifurcation of single fixed point value to multiple values is hardly visible, indication a very short bifurcation process from single fixed points to multiple points before transitioning to chaos. It is well-known that non-linear systems approach chaos exponentially fast at the same ratio of the Feigenbaum constant \cite{feigenbaum1975universality}. Therefore, the wider is the first bifurcation regime, generally the slower the system approaches chaos. 

That being said, the bifurcation process in high dimensional systems like LSTM may differ from that of the low dimensional $\tanh$ map or the logistic map, and more studies are needed to understand it better. Specifically, the low dimensional 1-D maps have much simpler and clear theoretical understanding of the multple chaos-order transitions: due to the prime number of periodicities; Such a phenomenon can appear in high dimensional systems like neural networks, yet the interactions among different dimensions may lead to much more complex transition processes, as seen in some of our other experiments in Appendixes Fig.~\ref{appendix:additional}.

To give a more detailed plausible theoretical explanation of the such multiple descents and order chaos transitions drawing parallels with the $\tanh$ map, we borrow the theoretical framework from Ref~\cite{16}. In Ref~\cite{16}, the authors have shown that for a generic high dimensional dynamical operator $f(\boldsymbol{x})$, the critical point that it transitions from order to chaos can be approximated by the following equation:
\begin{equation}
\frac{1}{N} \| \boldsymbol{J}^*_{\boldsymbol{f}} \|^2_F = 1
\label{eq:critical_point}
\end{equation}
where $\boldsymbol{J}^*_{\boldsymbol{f}}$ is the asymptotic Jacobian matrix of the dynamical operator $f$ at the critical point, and $\|\cdot\|_F$ is the Frobenius norm. 
Similar to 1-D maps, one can also get the critical point for not just $f$, but prime number of iterations of $f$, i.e., $f^2$, $f^3$, $f^5$, etc, i.e. $f^k$ where $k$ is a prime number. More explicitly, the critical point for $f^k$ can be approximated by the following equation:
\begin{equation}
\frac{1}{N} \| \boldsymbol{J}^*_{\boldsymbol{f}^k} \|^2_F = 1
\label{eq:critical_point_k}
\end{equation}
The critical points for $f^k$ with different prime number $k$ can be different, and they can be ordered in a way that the critical point for $f$ is the smallest, followed by $f^2$, $f^3$, etc. Therefore, as the model weights increase during training, the model can first enter into chaos at the critical point for $f$, then as the weights keep increasing, it can enter into chaos at the critical point for $f^2$, then $f^3$, etc, leading to multiple order chaos transitions similar to 1-D maps. Again, we have to caveat that in high dimensional systems like LSTM or neural networks in general, the interactions among different dimensions may also lead to further complex transition processes, not just limited to the prime number of periodicities as in low dimensional systems. Therefore, the above theoretical explanation is only a plausible one, and more studies are needed to understand the detailed transition processes in high dimensional systems like neural networks.

To ensure the reproducibility of the results, we have repeated the experiment with different random seeds, and the results consistently show the same phenomena of multiple descents and order chaos transitions. The most optimal epoch number may vary as seen in Appendixes Fig.~\ref{fig:sharp} and Movie S1 and S2, but it always coincides with the first order-to-chaos transition point. Additionally, the multiple descents are also always observed in the overfitting regime, with the locally optimal epoch (bottom of each loss cycle) coinciding with a transition between order and chaos. 

To further confirm the multiple descents phenomena, we carry out more extensive experiments on different datasets, optimizers and random seeds, and calculate the asymptotic distances, reduced sums and FTLEs for each as shown in the series of plots in Supplementary Information (SI). It can be seen that the multiple descents' phenomenon may not emerge in two conditions. 1. When the learning rate is very small, or the model size is small, the model may stay in the order phase during the whole training process. Without the phase transition between order and chaos, the multiple descents' phenomenon does not emerge. 2. When using SGD optimizer, the model moves very slowly in the parameter space, such that it is not able to travel far enough to cross multiple order chaos phases during the whole training process, therefore, the multiple descents' phenomenon does not emerge. However, in all of the SGD experiments, we can clearly see that when the test loss suddenly enters the new region of values and stay there, the FTLE also drastically changes sign and stay there too, meaning it enters into a new order/chaos phase. This further confirms the relationship between the test loss and the order/chaos transitions, even if the multiple descents' phenomenon does not emerge in some of the experiments.

One thing to note is that, the order chaos transition during the multiple descents are not always drastic as in Fig.~\ref{fig:lstm_chaos}(a). As seen in Fig.~\ref{fig:sharp}, the asymptotic distances can have less drastic change at the locally optimal epoch than that in the first experiment shown in Fig.~\ref{fig:lstm_chaos}(a). But nevertheless, they are still clearly transitions since the sudden change in asymptotic distances occur within one epoch that coincides with the sharp drop in test loss. The less drastic changes near the bottom of the cycle (which is also order to chaos transition point) is possibly due to one type of chaotic attractor to another, or the narrow order/chaos phase near the epoch as we illustrate in comparison with the well studied $\tanh$ map in Appendixes Fig.~\ref{appendix:additional}. Another phoenomena worth noting is that, in some of the experiements, like the one in Movie S2, the chaotic attractor in terms of the reduced sum $\boldsymbol h_{1599} \cdot \boldsymbol{1}$ converges slowly to narrow bands, and the asymptotic distances are also very small as seen in Fig.~\ref{fig:sharp}(b). This could potentially mean the chaotic phase is actually transient, which may eventually lead to order. However, within the number of iterations in our experiments, a clear convergence to order is not yet observed. The detailed investigation of these transitions in the overfitting phase requires further studies, and we leave it for future work.

To confirm the consistency of the multiple descents' phenomenon, we carry out repeated experiments on two typical settings learnt from the above experiments: 1. low learning rate and small model size, where the model is mostly in the order phase during the whole training process; 2. high learning rate and large model size, where the model is mostly in the chaotic phase during the whole training process. We repeat each experiment 10 times with different random seeds but otherwise identical settings, and count the number of descents in each experiment. The results are summarized in Table~\ref{tab:table2}, and the detailed descents plots are in Appendix Fig.~\ref{fig:statistics_positive} and Fig.~\ref{fig:statistics_negative}. Details of the two models are in~\ref{appendix:statistics}.

\begin{table}[hb]

\caption{Statistics of multiple descents in differet experiments with different learning rate and hidden size. Each set of experiment is repeated 10 times with different random seeds but otherwise identical settings. Details descents plots are in Appendix Fig.~\ref{fig:statistics_positive} and Fig.~\ref{fig:statistics_negative}.}
\label{tab:table2}
\centering
\begin{tabular}{|l|c|c|}
\hline
\textbf{Experiment} & Low LR and small size  & High LR and large size  \\
\hline
\textbf{Mean Descents count} & 0 & 15.8 \\
\hline
\textbf{Min Descents count} & 0 & 12\\
\hline
\textbf{Max Descents count} & 0 & 18\\ 
\hline
\end{tabular}
\end{table}

\section{Relationship between test loss and chaos during overfitting}
\label{sec:loss-distance}

\begin{figure}[H]
\centering
\includegraphics[width=\columnwidth]{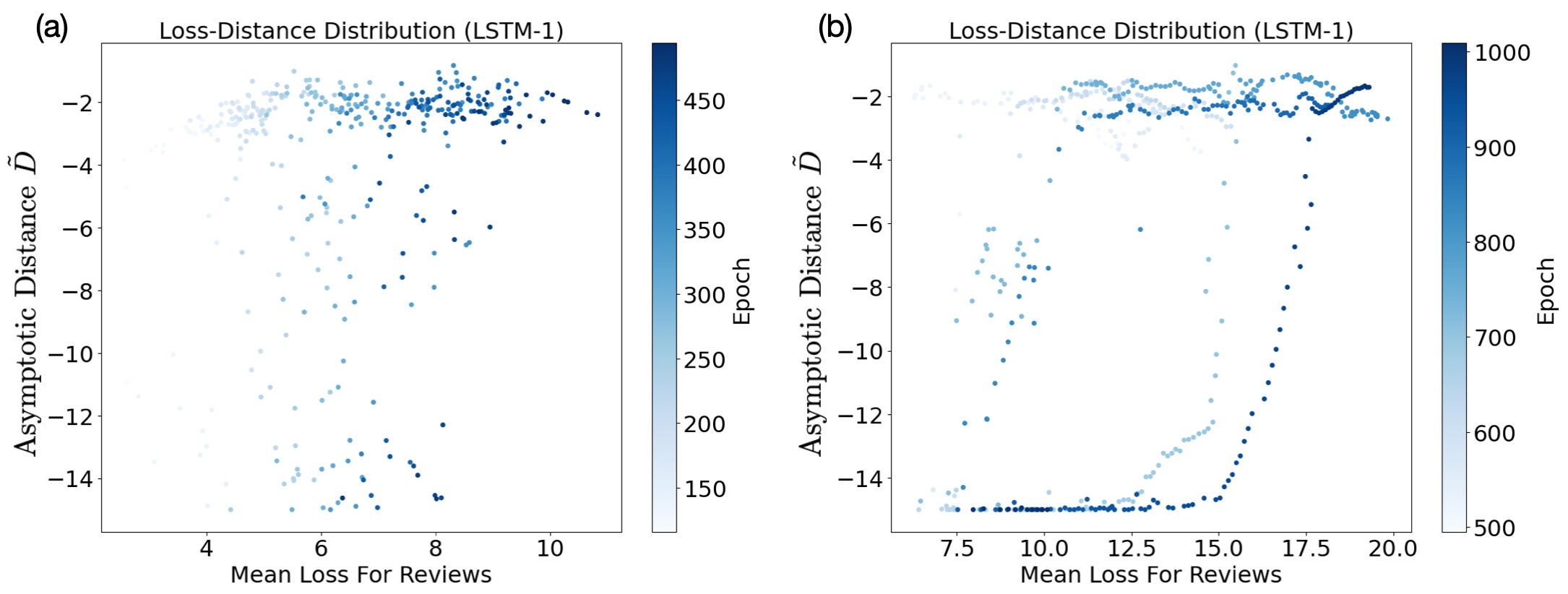}
\caption{{\bf Relationship between test loss of wrong predictions and chaos (as measured by asymptotic distance).} (a) illustrates the overfitting regime before clear multiple descents happen, i.e., between epoch 115 and 495. (b) illustrates the region where clear multiple descents occur, i.e., after epoch 495.}
\label{fig:loss-distance}
\end{figure}

The multiple descents occur more clearly in the overfitting regime in our experiments, or in other words after model has been well-trained. However, the large changes in loss function does not lead to similar changes in the model accuracy. This hints that while the model is less confident in its predictions as test loss decreases, it is not making significantly more incorrect predictions. Such phenomenon has been previously reported, that when overfitting, the increase in model loss can be attributed to an increase in loss on the incorrectly classified subset, rather than an increase in loss on the entire test set \cite{14}.

In our experiment from Section \ref{section:experiment}, we used 500 reviews to generate the results for asymptotic distances. Here, we use the same reviews but focus specifically on those with incorrect model predictions which is approximately 75 samples (the exact value varies between epochs and models as the system experiences slight fluctuations in accuracy during the training process). The loss is then calculated only on this reduced subset of reviews.

Plotting the asymptotic distances against the loss on this subset (Fig.~\ref{fig:loss-distance}) reveals a clear relationship between these quantities during each descent cycle. In Fig.~\ref{fig:loss-distance}(b), we observe a clear correlation between loss and distance, with increasing loss generally relating to increased asymptotic distances. Furthermore, in Fig.~\ref{fig:loss-distance}(b), the distances increase rapidly together with the test loss, then reach a plateau. Such a plateau in distances is due to the fact that the activation functions in the LSTM saturate, as they are hyperbolic activation functions. This is consistent with the behaviour of the $\tanh$ map in Fig.~\ref{fig:tanh_bifurcation}, where the system becomes less sensitive to changes in the input as it becomes very chaotic. Such a plateau does not mean the model is equally chaotic, as it can still reach the saturation distances at different rate. That is why the loss values still increases during this plateau. This is the same as the $\tanh$ map, where when the asymptotic distance reaches a plateau in Fig.~\ref{fig:tanh_bifurcation}, the Lyapunov exponent that characterize chaos actually increases.

Notice that before the multiple descents set in at epoch 495, as seen in Fig.~\ref{fig:loss-distance}(a), even if the model overfits after epoch 114, the monotonic relationship between test loss and chaos is not clear. This is likely because the model is mostly in the chaotic phase and the asymptotic distances are easily saturated. This can be seen in Fig.~\ref{fig:lstm_chaos}(b), where the 500 samples have a large dispersion between epoch 115 and 495 in most of the time, indicating that the model is mostly chaotic. One thing to note between epoch 114 and 495 is that, there are narrow regimes of order phase in between this wide and mostly chaotic regime. This can be seen from Fig.~\ref{fig:lstm_chaos}(a) when the asymptotic distances are around -15, and also from Fig.~\ref{fig:lstm_chaos}(b) when the reduced sum of the 500 samples converge to one unique value, both very briefly. The order phases are so narrow, that on the whole scale of things the associated changes in test loss appears as noise. But in actual fact, they are intermittent order phases that are associated with the local minima in the loss function. Similar narrow gaps of order can also be seen in the $\tanh$ map in Fig.~\ref{fig:tanh_bifurcation}, and it has been well established in non-linear dynamics that these order phases after the first transition to chaos arrives from period-doubling bifurcation of prime-numbered periodic cycles \cite{sander2011period}.

\section{Discussion}

In this work, we discovered a novel feature of `multiple descents' during the training process of a deep learning model. Our findings suggest a close empirical association between these multiple descents and the intrinsic dynamical transitions between order and chaos within the network. By leveraging asymptotic stability analysis, akin to the approaches proposed by \cite{16} to explore optimal machine intelligence at the edge of chaos. However, unlike their work which deems the asymptotic edge of chaos as the optimal model configurations, we find that in our experiments only the first edge of chaos appears to be the most optimal, whereas the others are simply local optimal. While we explored this in the various experimental settings tested, it is yet to be more extensively validated on other network architectures. Additionally, alternative contributions from optimization dynamics or hyperparameter choices may also play a role in the phenomena, and yet to be explored.  This may provide a new perspective on when and why neural networks achieve peak performance, differing from previous studies which might focus on model size, dataset characteristics, or learning rate adjustments \cite{Mei2022, Smith2017}. Our work not only extends the theoretical framework beyond the traditional bias-variance trade-off but also offers practical implications for model training, potentially suggesting new strategies for training models at optimal epochs and possibly enhancing model generalization by understanding and navigating these chaotic dynamics with more principles. This could in turn lead to more robust deep learning models, especially in scenarios where traditional overfitting mitigation strategies fall short \cite{Goodfellow2016, Neyshabur2017}.

It is worth noting that the multiple descents phenomenon we observed in LSTM could share some connection with the `edge of stability' phenomena observed in Ref~\cite{Cohen2021}. The plausible connection is that both edge of chaos in the model and edge of stability in the loss functions' Hessian affect the local loss landscape. However, their direct connection may not be straightforward, as the edge of stability is defined relative the learning rate of the optimization algorithm, while the edge of chaos is defined in terms of the asymptotic stability of the model against perturbation to the input values, and is largely intrinsic to the model. However, it could still be possible that the former is still mostly model dependent as the Hessian is evaluated on the full-batch as in Ref~\cite{Cohen2021}. The relationship between these two concepts is an interesting avenue for future research, and it may require a deeper theoretical understanding of the dynamics of neural networks and their loss landscapes.

One particular intersting finding of our work is that, despite its high dimensions, the LSTM training process exhibits a phenomenologically similar phase diagram structure as the well-known $\tanh$ map in dynamical systems, that the systems goes through bifurcation process to chaos, with the first order chaos transition being the widest, followed by many narrower regimes of order phases. Given the generic nature of this bifurcation diagram structure across different non-linear dynamics other than hyperbolic functions, we conjecture, with some theoretical parallel to 1-D maps, that the qualitative patterns in our findings are not unique to LSTM, but may also generalize to other deep learning models. This opens up a new avenue of research in understanding the training dynamics of deep learning models, leveraging on the established theories in non-linear dynamical systems, and potentially lead to new training strategies that can improve the generalization and robustness of the model. That being said, more in-depth theoretical studies are needed to understand the exact reason behind the similarities between LSTM and the $\tanh$ map, and to explore the generalizability of our findings to other deep learning models.

Another interesting unanswered questions is whether this multiple descents phenomenon is connected to the well-known `double descent' phenomenon in deep learning, in particular epoch-wise double descent\cite{Nakkiran2021}. Recent work has also observed double descent in different contexts such as deep reinforcement learning \cite{vesely2025presence}, suggesting that this phenomenon may be more general than initially thought. Apart from using different network models, there are two key differences between the two phenomena. First, the double descent phenomena require label noise in the training data to achieve, whereas the multiple descents phenomena do not. Adding label noise may force the model to become chaotic in order to overfit the noisy labels, which could potentially be the cause of the `ascent' of test loss similar to our findings. Second, the double descent phenomenon shows a smooth change of descent, indicating no abrupt phase transition like ours. A more detailed investigation using the same experimental setup is needed to clearly understand the relationship between the two phenomena, and whether they are related to each other.

A limitation to this finding is that, similar analysis on feed-forward neural networks like multi-layer perceptrons and convolutional neural networks did not show clear multiple descents phenomena before, although the first order to chaos transition coinciding with optimal performance was observed~\cite{zhang2024asymptotic,16}. One possibility is that the multiple descents phenomena may only appear in certain types of tasks or datasets, and further studies are needed to explore when and why such phenomena occur. Nevertheless, our findings in LSTM provide a new perspective on the training dynamics of deep learning models, and open up new avenues for future research in this area.

%\section*{Acknowledgements}

% Acknowledgements can be included here if applicable.

\section*{Data availability}

The code of this study is available at \url{https://github.com/WEEMUNPARK/nus-lstm-dynamic-behaviors-publication} and \url{https://github.com/xufannorman/LSTM_EOC}.

\section*{Declaration of generative AI and AI-assisted technologies in the writing process}
During the preparation of this work the author(s) used XAI/Grok and Anthropic/Claude in improving the presentation and literature review. After using this tool/service, the author(s) reviewed and edited the content as needed and take(s) full responsibility for the content of the published article.

\appendix

\section{Additional LSTM models}
\label{appendix:additional}

\begin{figure}[H]
\centering
\subfloat[]{\includegraphics[width=\columnwidth]{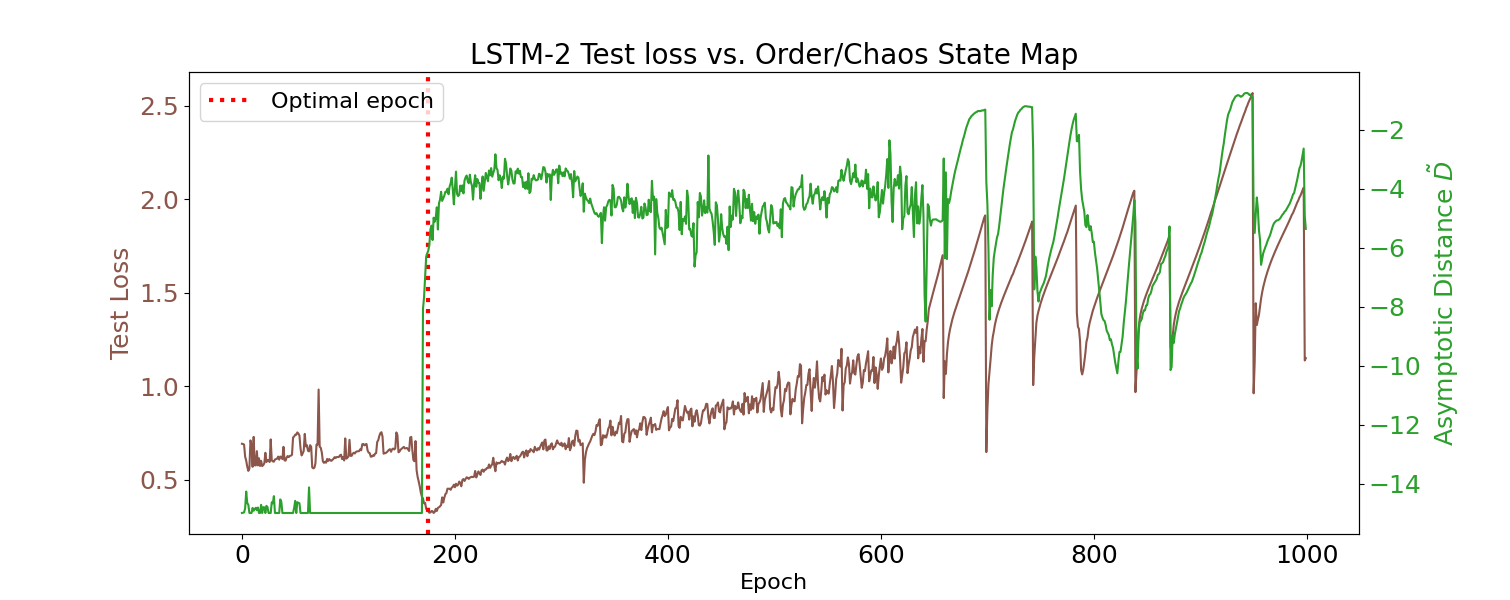}}
\\
\subfloat[]{\includegraphics[width=\columnwidth]{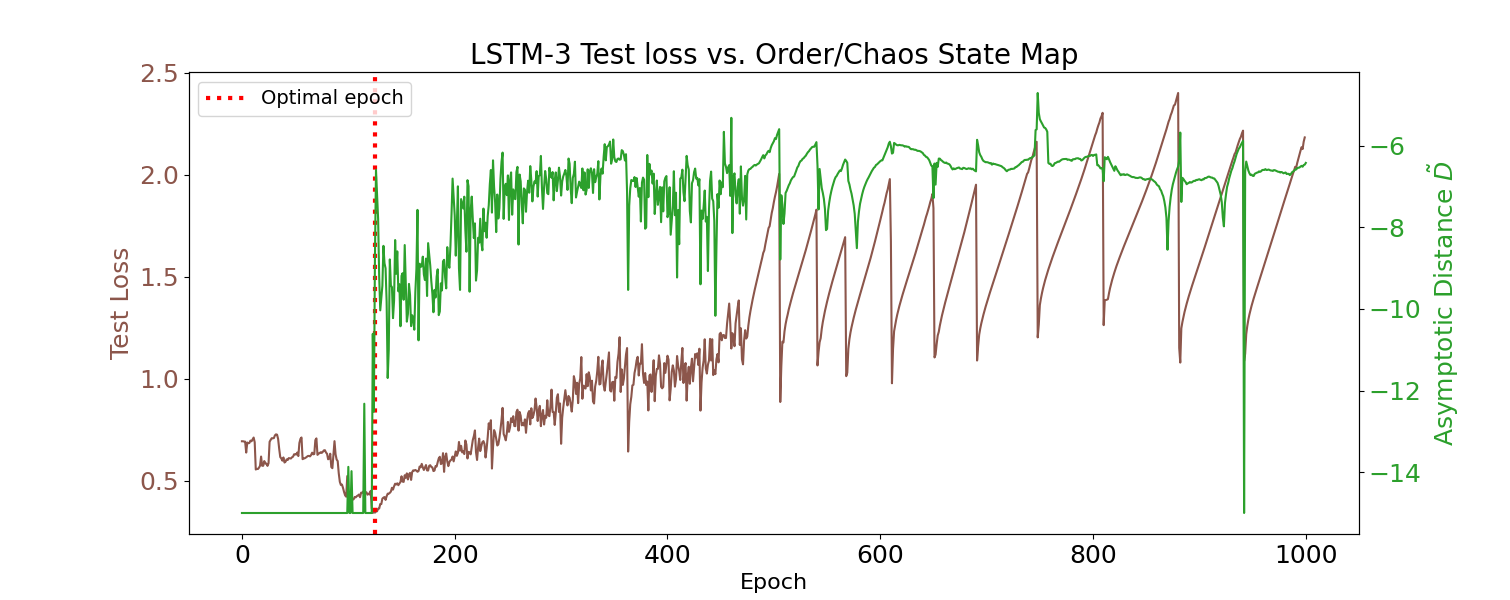}}
\caption{Two other experiments with the same setting showing multiple descents, and best epoch occurring at the first order to chaos transition at (a) epoch 190 and (b) epoch 120.}
\label{fig:sharp}
\end{figure}

Fig.~\ref{fig:sharp} shows two other experiments that have slightly different look from results in Fig.~\ref{fig:lstm_chaos}(a), with different random seeds. Note that we use the model in the main text because these two experiments have slightly worse performance in terms of best test accuracy than the model presented in the main text. Unlike in Fig.~\ref{fig:lstm_chaos}, the asymptotic distances in Fig.~\ref{fig:loss-distance} do not drop to 0 (or $-15$ in our case) at the end of each cycle, but to moderate values around $-8$. This is likely due to the fact that the order/chaos transition is very narrow around the locally optimal epoch, and the resolution of our sampling (1 epoch) is not enough to capture the full picture of the order chaos transition. In other words, the order chaos transitions still exist, but they lie between two adjacent epochs. Since our measurements do not afford resolution higher than every epoch, this misses the exact transition point where the asymptotic distances should be 0.

\begin{figure}[ht]
\centering
\subfloat[]{\includegraphics[width=0.5\columnwidth]{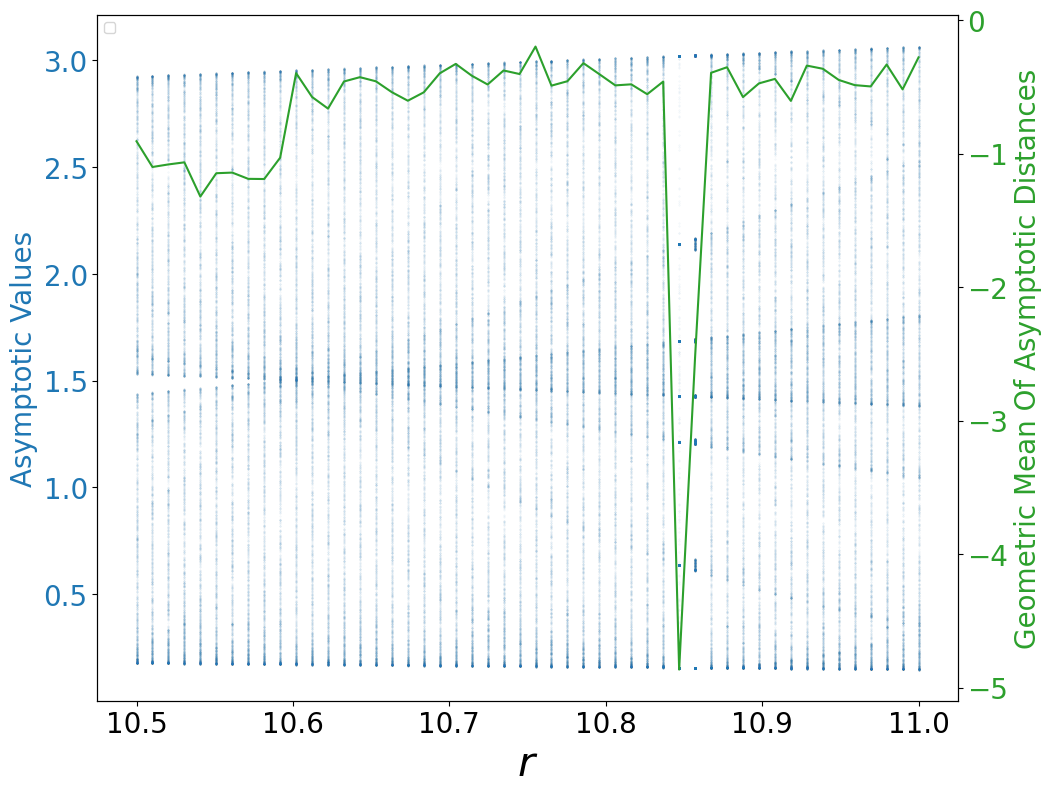}}
\hfill
\subfloat[]{\includegraphics[width=0.5\columnwidth]{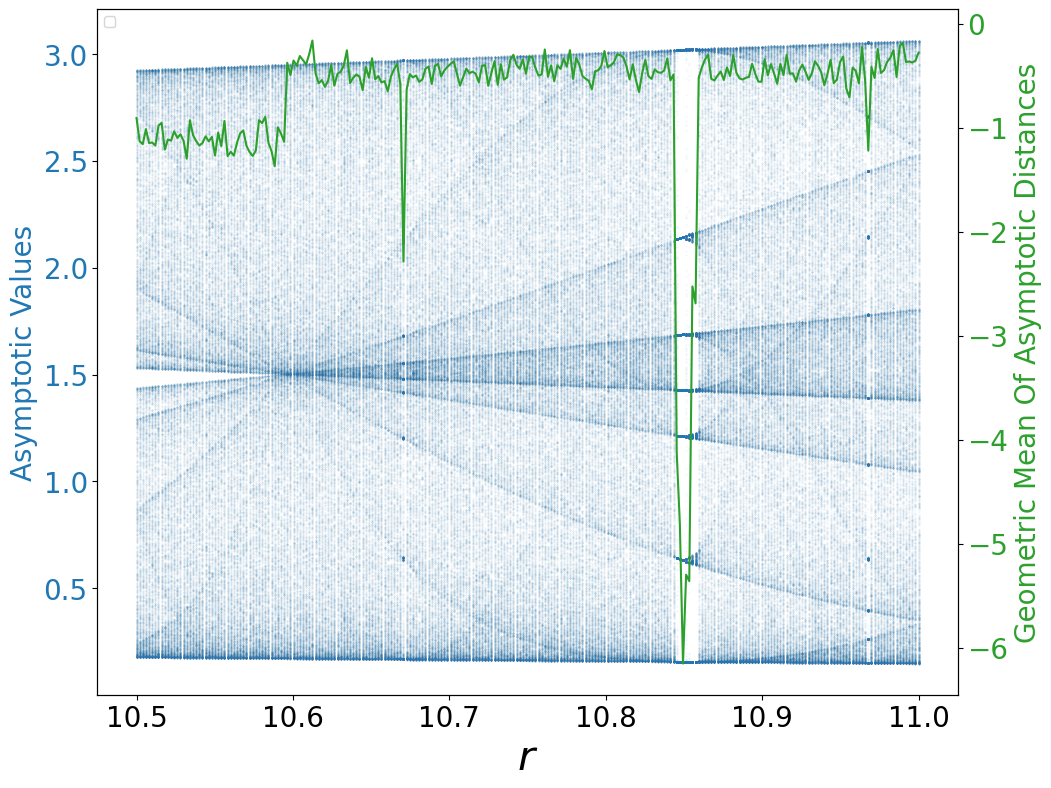}}
\caption{$\tanh$ map plotted at two different sampling resolutions, with $r$ ranging from 10.5 to 11. The low resolution map (a) covers 50 different values of $r$, while the high resolution map (b) covers 281 different values of $r$. The high resolution map (b) clearly shows the transitions to order at $r=10.675$ due to the presence of an order phase, while the low resolution map (a) misses this transition.}
\label{fig:tanh-aliasing}
\end{figure}

This is further illustrated in Fig.~\ref{fig:tanh-aliasing} on the $\tanh$ map, which we previously drew analogy to. In high resolution map Fig.~\ref{fig:tanh-aliasing}(b) with 281 different $r$ values, we can clearly see a very narrow regime of order phase around $r=10.675$, where the asymptotic distance drops drastically. However, in the low resolution map Fig.~\ref{fig:tanh-aliasing}(a) with only 50 different $r$ values, the order phase is not captured, and the asymptotic distance stays high at $r=10.675$.

\section{FTLE robustness}

\begin{figure}[H]
\centering
\subfloat[]{\includegraphics[width=.9\columnwidth]{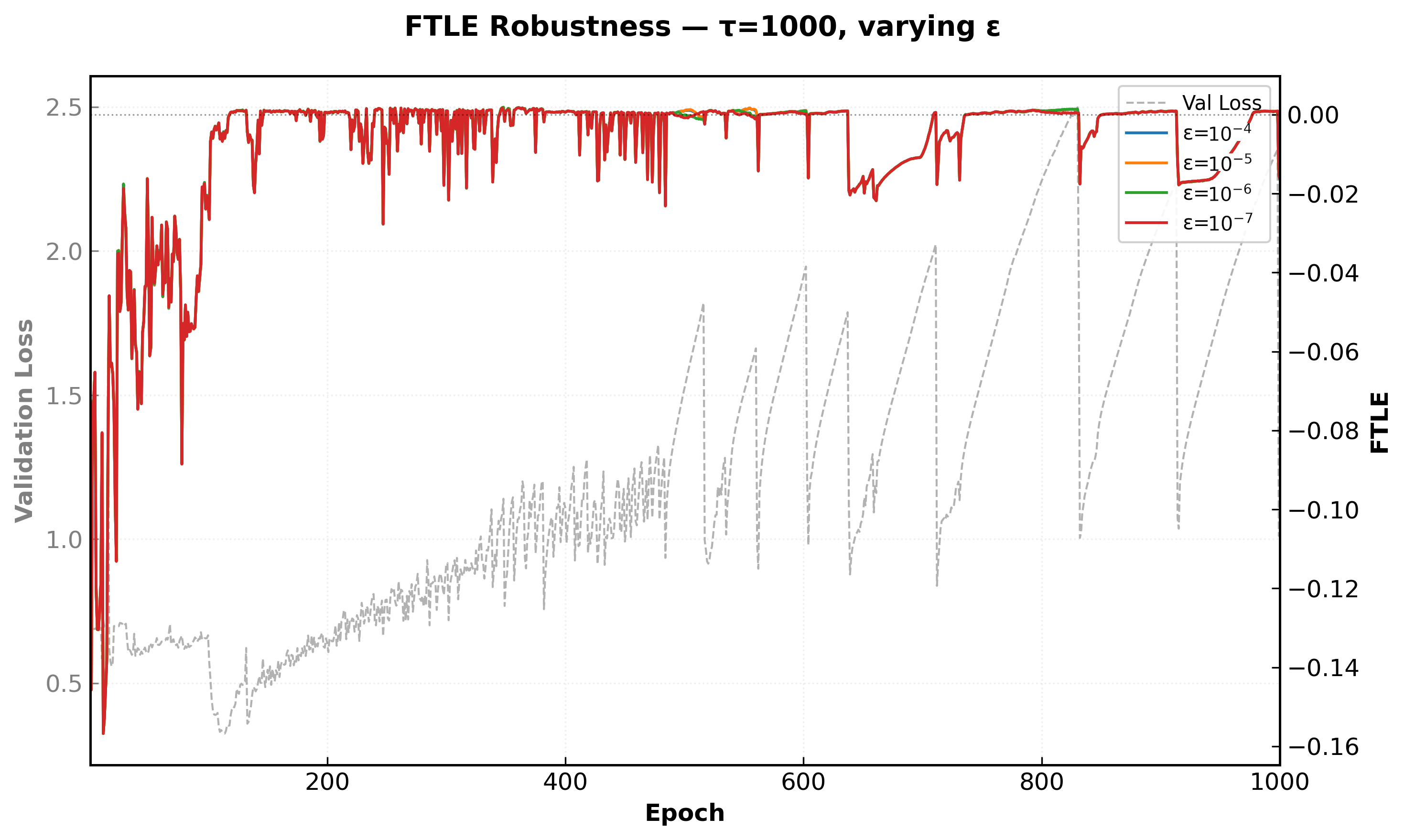}}
\\
\subfloat[]{\includegraphics[width=.9\columnwidth]{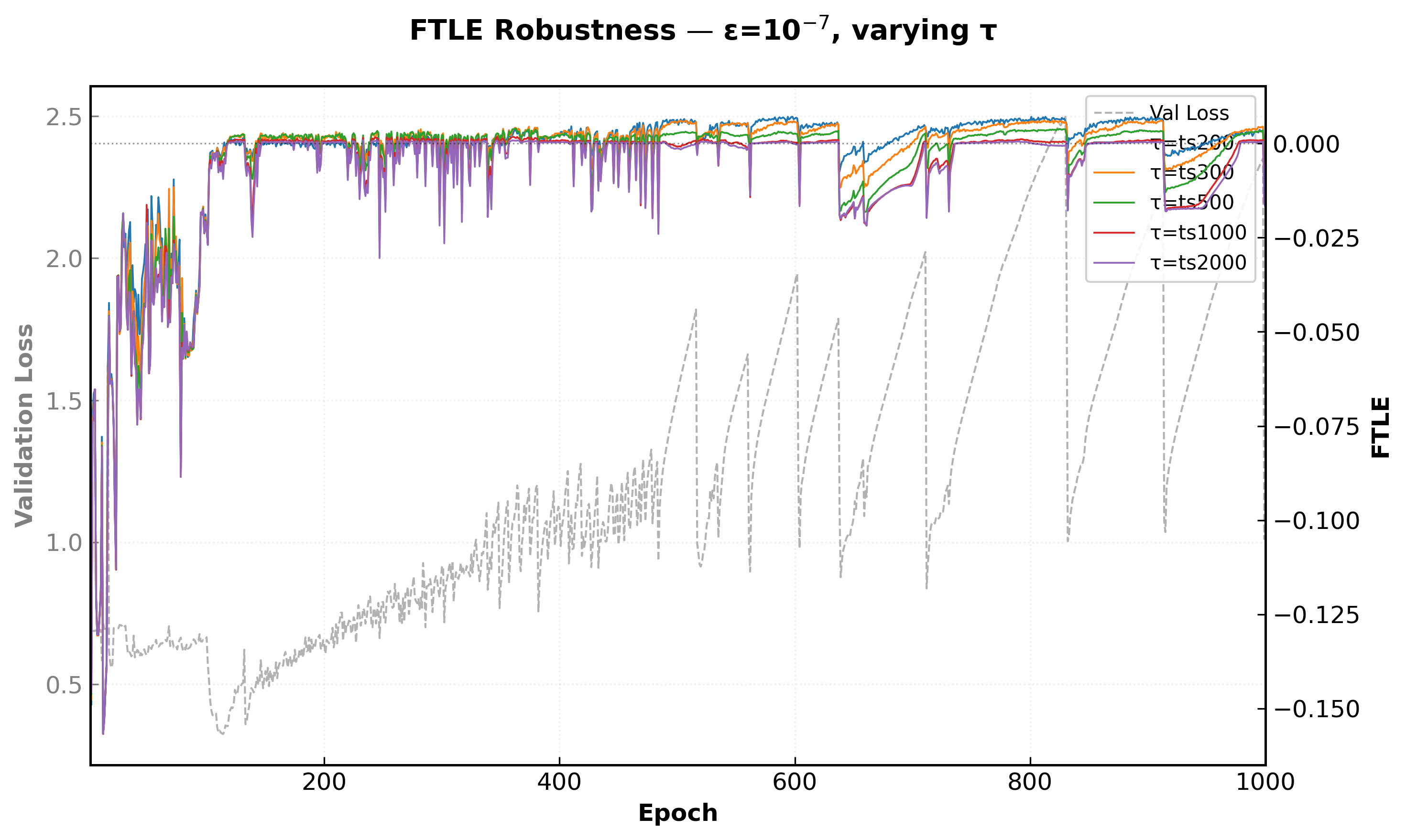}}
\caption{Robustness of the finite time Lyapunov exponent (FTLE) results to different values of noise level $\varepsilon$ at $10^{-7}$, $10^{-6}$, $10^{-5}$ and $10^{-4}$. (a) and time horizon $\tau$ at 200 to 2000. (b). The overall trends of FTLE and the transition points between order and chaos remain consistent across different settings, indicating the robustness of our findings regarding the order-chaos transitions during the training process of LSTM.}
\label{fig:ftle-robustness}
\end{figure}

In estimating the finite time Lyapunov exponent (FTLE) using the Benettin estimator~\cite{Benettin1980a}, there are two key parameters to set that may affect the results: the noise level $\varepsilon$ and the time horizon $\tau$. In Fit.~\ref{fig:lstm_chaos}, we use $\varepsilon=10^{-7}$ and $\tau=1000$. Here we have tested the robustness of the FTLE results to different values of these two parameters, and the results are shown in Fig.~\ref{fig:ftle-robustness}. It can be seen that the overall trends of FTLE are robust to different values of $\varepsilon$ and $\tau$, with the transition points between order and chaos remaining consistent across different settings. This indicates that our findings regarding the order-chaos transitions during the training process of LSTM are not sensitive to the specific choices of these parameters, further strengthening the validity of our conclusions. Note that when FTLE is near 0, i.e. the order-chaos transition point, the FTLE values can be more sensitive to the choice of $\varepsilon$, typical of the nature of the transition point. However, the transitions from positive to negative FTLE are consistently at the sudden descent epochs across different $\varepsilon$ values, further confirming the relationship between the order-chaos transitions and the sudden test loss descents.

\section{Statistical robustnessof the multiple descents}
\label{appendix:statistics}

\begin{figure}[H]
\centering
\includegraphics[width=.8\columnwidth]{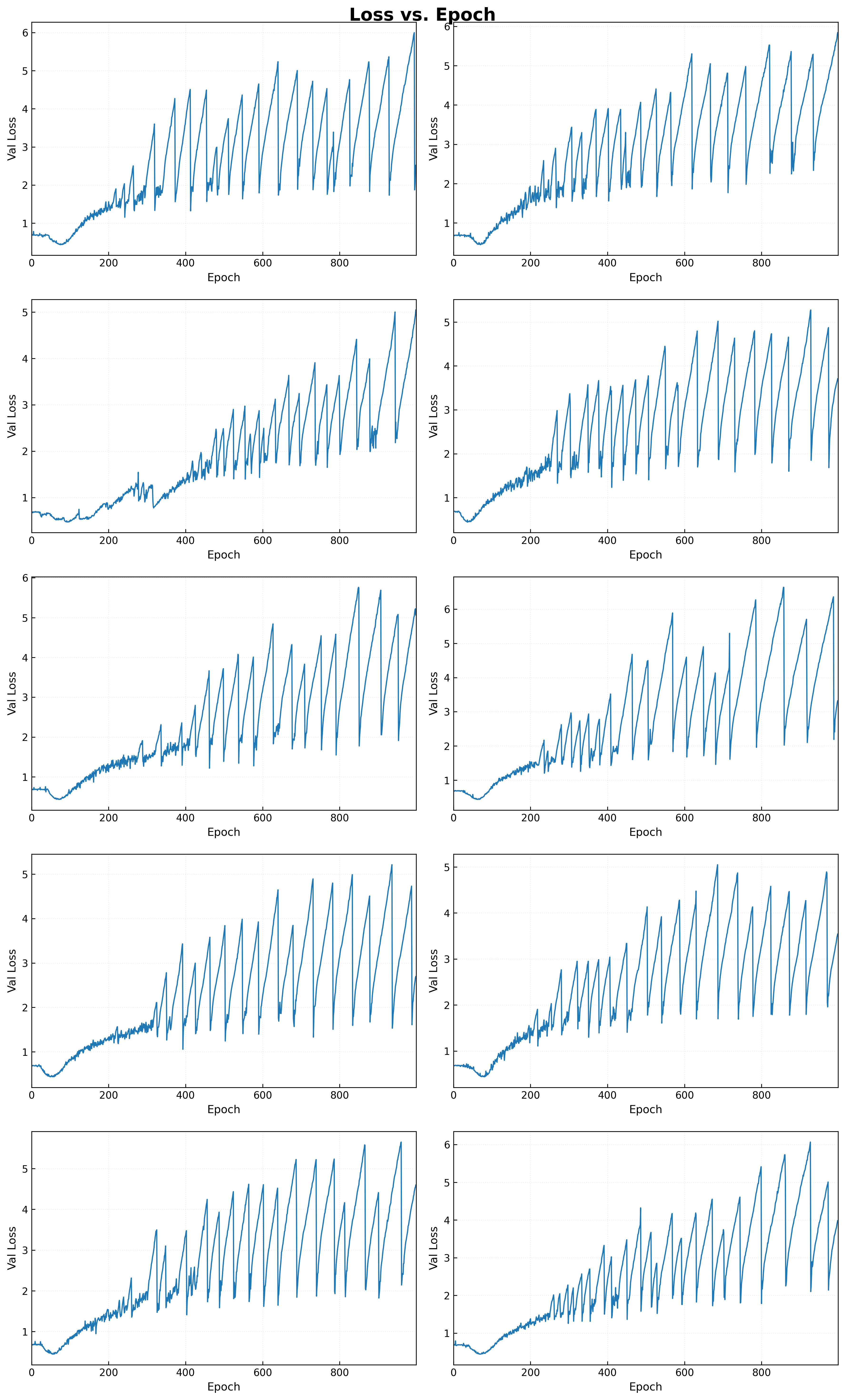}
\caption{{\bf Robustness analysis of the multiple descents when they show up.} All of the 10 experiments with the same setting but different random seeds show the multiple descents phenomenon. Learning rate is set at 0.0005, Adam optimizer, and hidden dimension 100.} 
\label{fig:statistics_positive}
\end{figure}

\begin{figure}[H]
\centering
\includegraphics[width=.8\columnwidth]{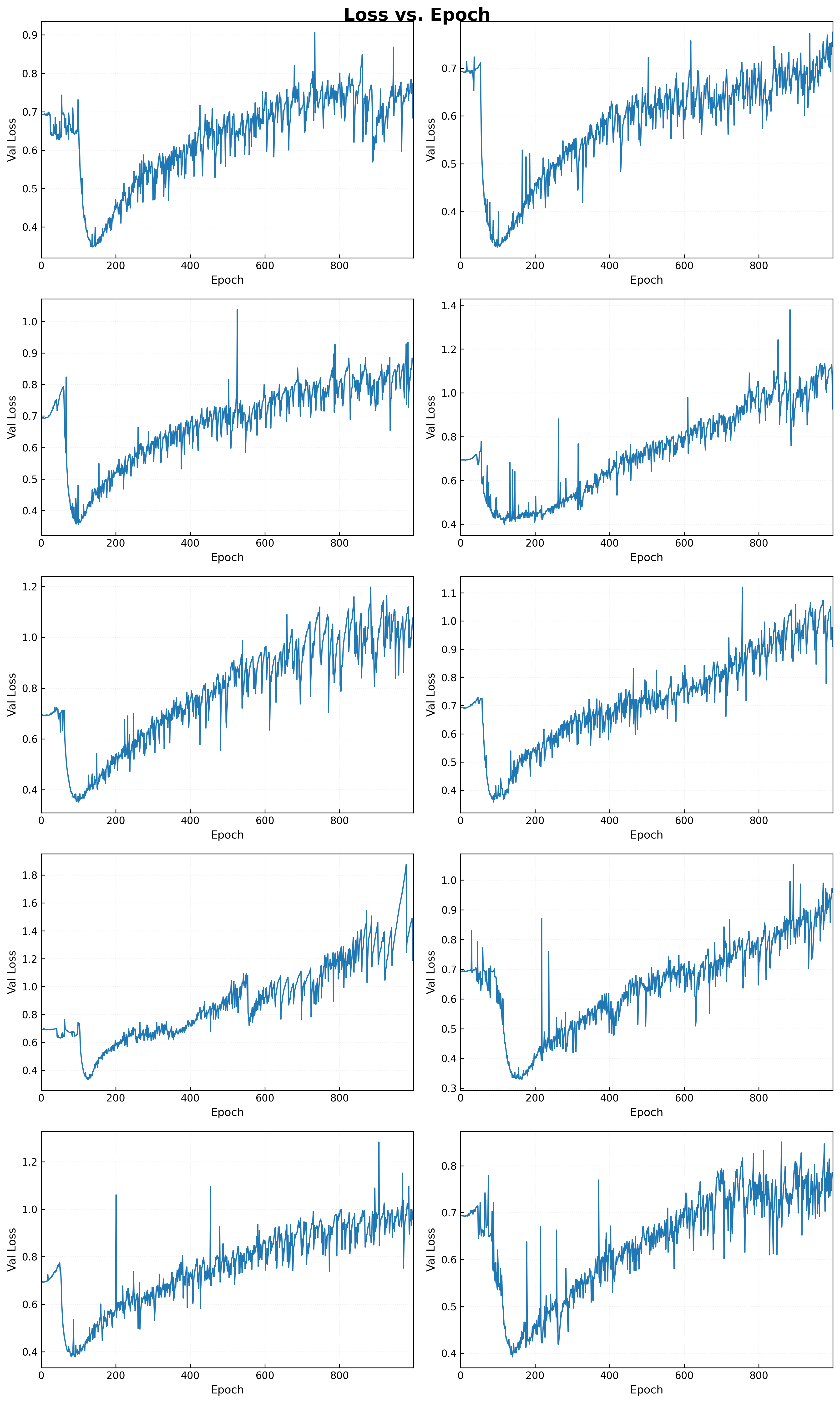}
\caption{{\bf Robustness analysis of the multiple descents when they do not show up.} All of the 10 experiments with the same setting but different random seeds show the multiple descents phenomenon. Learning rate is set at 0.0003, Adam optimizer, and hidden dimension 60.} 
\label{fig:statistics_negative}
\end{figure}

To ensure the reproducibility of the results, we have repeated the experiment with different random seeds, and the results consistently show the same phenomena of multiple descents or the lack of it. The two different model setups are chosen in such a way that one has higher learning rate and larger model size, which leads to more chaotic dynamics and multiple descents, while the other has lower learning rate and smaller model size, which leads to more ordered dynamics and no multiple descents. If intermediate settings are chosen, the results can be more mixed, and less useful to indicate the trends and robustness of the multiple descents phenomenon. Specifically, the first model has learning rate at 0.0005, Adam optimizer, and hidden dimension 100, while the second model has learning rate at 0.0003, Adam optimizer, and hidden dimension 60, both are trained on the Yelp review dataset. The results of the two models are shown in Fig.~\ref{fig:statistics_positive} and Fig.~\ref{fig:statistics_negative} respectively. To count the number of descents, we define a descent as a drop in test loss that is at least 1 (a very significant drop in just a single epoch), and preceded by 20 epochs of continuous increase in test loss. The statistics of the descents count in the two models are summarized in Table~\ref{tab:table2}.

\bibliographystyle{unsrt}
\bibliography{Untitled}

%% -------------------------------------------------------
%% Supplementary Information title page
%% -------------------------------------------------------
\newpage
\thispagestyle{plain}
\vspace*{\fill}
\begin{center}
  {\Huge\textbf{Supplementary Information}}\\[1.5em]
  {\large\textit{Multiple Descents in Deep Learning as a Sequence of\\[0.3em]
  Order-Chaos Transitions in LSTM Networks}}
\end{center}
\vspace{2em}
\noindent The following figures present additional experimental results across a wider range of hyperparameter configurations, datasets (IMDB and YELP), and optimizers (Adam and SGD), further supporting the main findings reported in the paper. Figures S1--S4 show results on the IMDB dataset with the Adam optimizer at different learning rates and hidden dimensions; Figures S5--S9 show results on the YELP dataset with the Adam optimizer; Figures S10--S14 show results on the IMDB dataset with the SGD optimizer. In each figure, panel (a) shows the test loss together with the asymptotic distance $D$, panel (b) shows the test loss together with the reduced sum of the recurrent output $\boldsymbol{h}_t$, and panel (c) shows the test loss together with the finite-time Lyapunov exponent (FTLE). The dashed red vertical line marks the optimal epoch identified by the lowest test loss.
\vspace*{\fill}
\clearpage

%% Supplementary figures — each page of Additional_Experiments.pdf at text width
\begin{figure}[H]\centering
\includegraphics[page=1,width=\textwidth]{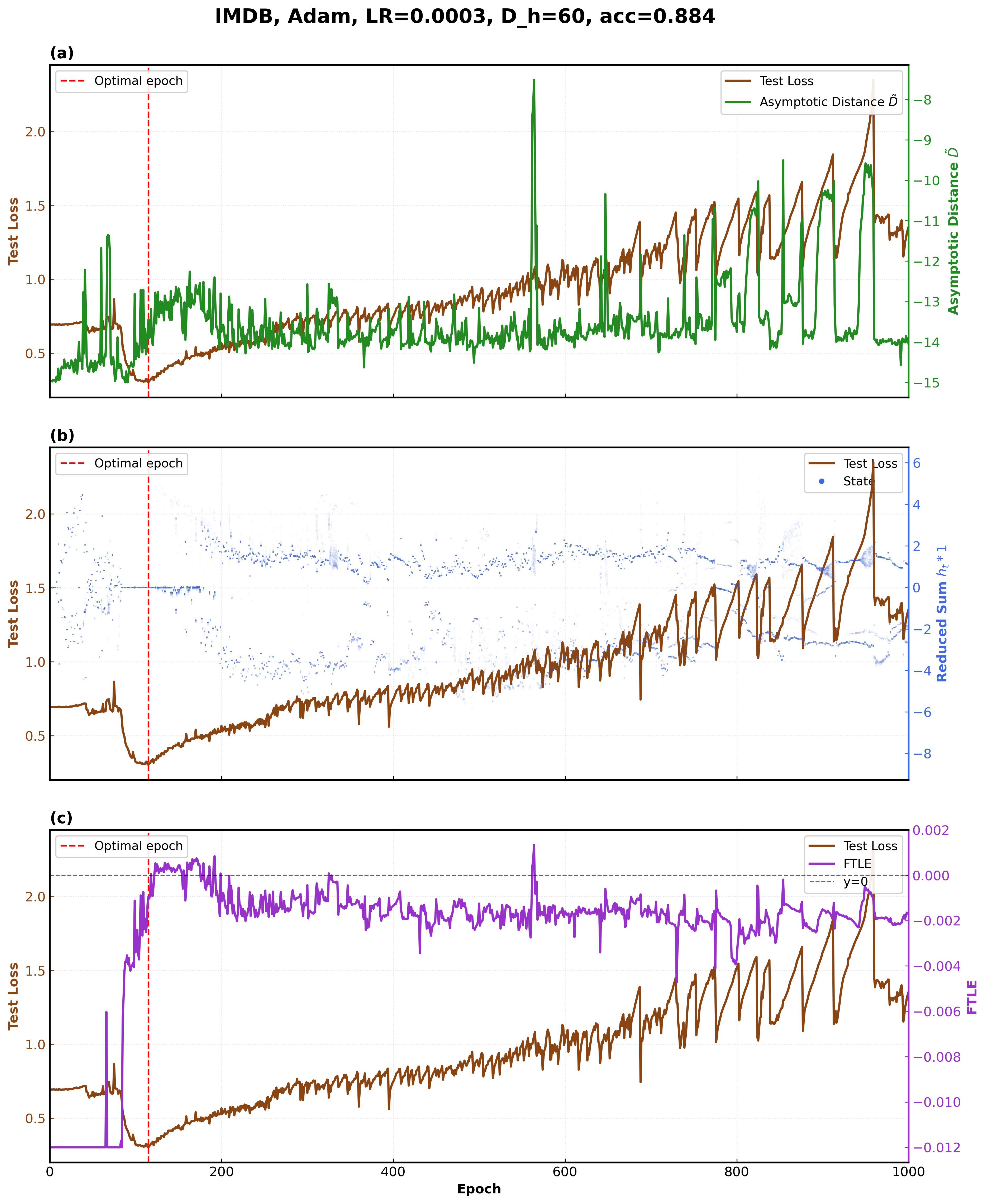}
\caption*{\textbf{Figure S1.} IMDB, Adam, LR\,=\,0.0003, $D_h$\,=\,60, acc\,=\,0.884.}
\end{figure}\clearpage

\begin{figure}[H]\centering
\includegraphics[page=2,width=\textwidth]{Additional_Experiments.pdf}
\caption*{\textbf{Figure S2.} IMDB, Adam, LR\,=\,0.0003, $D_h$\,=\,60, acc\,=\,0.867.}
\end{figure}\clearpage

\begin{figure}[H]\centering
\includegraphics[page=3,width=\textwidth]{Additional_Experiments.pdf}
\caption*{\textbf{Figure S3.} IMDB, Adam, LR\,=\,0.0003, $D_h$\,=\,80, acc\,=\,0.877.}
\end{figure}\clearpage

\begin{figure}[H]\centering
\includegraphics[page=4,width=\textwidth]{Additional_Experiments.pdf}
\caption*{\textbf{Figure S4.} IMDB, Adam, LR\,=\,0.001, $D_h$\,=\,100, acc\,=\,0.874.}
\end{figure}\clearpage

\begin{figure}[H]\centering
\includegraphics[page=5,width=\textwidth]{Additional_Experiments.pdf}
\caption*{\textbf{Figure S5.} YELP, Adam, LR\,=\,0.0003, $D_h$\,=\,60, acc\,=\,0.793.}
\end{figure}\clearpage

\begin{figure}[H]\centering
\includegraphics[page=6,width=\textwidth]{Additional_Experiments.pdf}
\caption*{\textbf{Figure S6.} YELP, Adam, LR\,=\,0.0003, $D_h$\,=\,60, acc\,=\,0.800.}
\end{figure}\clearpage

\begin{figure}[H]\centering
\includegraphics[page=7,width=\textwidth]{Additional_Experiments.pdf}
\caption*{\textbf{Figure S7.} YELP, Adam, LR\,=\,0.0001, $D_h$\,=\,80, acc\,=\,0.793.}
\end{figure}\clearpage

\begin{figure}[H]\centering
\includegraphics[page=8,width=\textwidth]{Additional_Experiments.pdf}
\caption*{\textbf{Figure S8.} YELP, Adam, LR\,=\,0.0005, $D_h$\,=\,80, acc\,=\,0.814.}
\end{figure}\clearpage

\begin{figure}[H]\centering
\includegraphics[page=9,width=\textwidth]{Additional_Experiments.pdf}
\caption*{\textbf{Figure S9.} YELP, Adam, LR\,=\,0.0005, $D_h$\,=\,100, acc\,=\,0.807.}
\end{figure}\clearpage

\begin{figure}[H]\centering
\includegraphics[page=10,width=\textwidth]{Additional_Experiments.pdf}
\caption*{\textbf{Figure S10.} IMDB, SGD, LR\,=\,1.0, $D_h$\,=\,60, acc\,=\,0.872.}
\end{figure}\clearpage

\begin{figure}[H]\centering
\includegraphics[page=11,width=\textwidth]{Additional_Experiments.pdf}
\caption*{\textbf{Figure S11.} IMDB, SGD, LR\,=\,1.5, $D_h$\,=\,60, acc\,=\,0.850.}
\end{figure}\clearpage

\begin{figure}[H]\centering
\includegraphics[page=12,width=\textwidth]{Additional_Experiments.pdf}
\caption*{\textbf{Figure S12.} IMDB, SGD, LR\,=\,2.0, $D_h$\,=\,60, acc\,=\,0.872.}
\end{figure}\clearpage

\begin{figure}[H]\centering
\includegraphics[page=13,width=\textwidth]{Additional_Experiments.pdf}
\caption*{\textbf{Figure S13.} IMDB, SGD, LR\,=\,2.0, $D_h$\,=\,60, acc\,=\,0.854.}
\end{figure}\clearpage

\begin{figure}[H]\centering
\includegraphics[page=14,width=\textwidth]{Additional_Experiments.pdf}
\caption*{\textbf{Figure S14.} IMDB, SGD, LR\,=\,2.0, $D_h$\,=\,80, acc\,=\,0.871.}
\end{figure}

\end{document}